\documentclass[runningheads]{llncs}

\usepackage{eccv}

\usepackage{eccvabbrv}

\usepackage{graphicx}
\usepackage{booktabs}
\usepackage{hyperref}
\usepackage{tabularx}
\usepackage[flushleft]{threeparttable}
\usepackage{multirow} 

\usepackage[accsupp]{axessibility}  

\usepackage{hyperref}

\usepackage{orcidlink}

\begin{document}

\title{Rethinking Pre-Training and Augmentation for Zero-Shot Cross-City Object Detection}

\titlerunning{Rethinking Pre-Training and Augmentation for ZCOD}

\author{
  Long Hoang Pham*\orcidlink{0000-0002-3240-657X} \and
  Quoc Pham-Nam Ho*\orcidlink{0009-0006-7256-4798} \and
  Huy-Hung Nguyen*\orcidlink{0000-0001-5394-9381} \and \\
  Duong Nguyen-Ngoc Tran\orcidlink{0000-0001-7537-6377} \and
  Ngoc Doan-Minh Huynh\orcidlink{0000-0003-0662-4036} \and
  Cu Quoc Le\orcidlink{0009-0005-9617-1983} \and \\
  Hoang Khang Nguyen\orcidlink{0000-0003-3616-0367}\and
  Hyung-Min Jeon \and
  Chi Dai Tran\orcidlink{0000-0001-6345-8658} \and
  Son Hong Phan\orcidlink{0009-0000-0572-6685} \and \\
  Duong Khac Vu\orcidlink{0009-0009-4715-0704} \and
  Trinh Le Ba Khanh\orcidlink{0000-0003-1688-9003} \and
  Jae Wook Jeon$\dagger$\orcidlink{0000-0003-0037-112X}
}

\authorrunning{L.~H.~Pham et al.}

\institute{
  \href{https://micro.skku.ac.kr/micro/index.do}{Automation Lab},\\
  Department of Electrical and Computer Engineering,\\
  Sungkyunkwan University, South Korea\\
  \email{\{phlong,hpnquoc,huyhung91,jwjeon\}@skku.edu} 
}

\maketitle


\begin{abstract}
  Real-world deployment of traffic surveillance systems is bottlenecked by geographic domain shift, in which models trained in one city underperform when applied to an unseen target city. Conventional domain adaptation relies on hyperparameter-sensitive architectures or direct profiling of target data. Both are fundamentally precluded in privacy-conscious ecosystems that require completely blind training and evaluation loops. In this setting, we explore the effects of pre-training and augmentation in addressing the domain shift problem. Specifically, we propose a new modular training pipeline for object detection structured around two core orthogonal pillars: (1) a multi-dataset pre-training strategy featuring a class-agnostic objectness distillation to decouple structural vehicle geometry from semantic taxonomies, and (2) a domain-resilient augmentation stream featuring a novel Grayworld transformation that forces global attention heads to strip volatile chromatic shortcuts in favor of robust shape priors. When evaluated with the real-time transformer-based detector RF-DETR, our framework bridges cross-city distribution gaps while using limited GPU memory (16GB). Our optimized variants, RF-DETR-HR and RF-DETR-Grayworld, deliver a substantial empirical gain of +24.29 over the baseline, achieving 1st place (47.53 mAP) on the AI City Challenge Track 6 leaderboard. Code and data are available at: \href{https://github.com/SKKUAutoLab/aic26_cross_city}{SKKUAutoLab/aic26\_cross\_city}.
\end{abstract}

\section{Introduction}
\label{sec:intro}
Real-world deployment of traffic surveillance systems (TSS) faces severe challenges from geographic domain shift. Object detectors perform well when trained and tested in the same location, but their accuracy drops sharply when applied to an unseen city with different road layouts, camera viewpoints, vehicle distributions, and environmental conditions \cite{Saunier2018, Zhou2026}. Despite the importance of this problem, robust cross-city generalization remains underexplored because large-scale surveillance data collection is constrained by privacy concerns.

To address this research gap, the 10th AI City Challenge (AIC2026)~\cite{AIC2026} Track 6 introduces a privacy-preserving benchmark using the Milestone Systems' Hafnia Training-as-a-Service (TaaS) platform~\cite{Hafnia}. In this setup, neither the source-city training data nor the target-city data can be accessed. Instead, methods must be containerized and run in fully blind training and inference loops. This setting requires practical, robust techniques rather than architecture scaling~\cite{YOLO26, DETR, Deformable-DETR, Co-DETR, RF-DETR} or direct profiling of target data~\cite{Tarvainen2017, Chen2018, TentFT, Chang2019, Le2024}.

In this work, we develop a robust, modular training pipeline for Zero-Shot Cross-City Object Detection (ZCOD). To address geographic domain shift without heavy parameter overhead, our framework focuses on two key stages: data preparation and multi-stage optimization. First, we introduce a multi-dataset pre-training strategy governed by a class-agnostic objectness distillation pass. By collapsing heterogeneous label taxonomies into a single binary foreground representation, the model extracts rich geometric inductive biases about vehicle structural boundaries and avoids class alignment conflicts across domains. Second, we propose a domain-resilient chromatic augmentation stream called Grayworld augmentation. This method forces global self-attention heads to remove volatile, sensor-dependent photometric shortcuts and favor robust, shape-centric representation learning. Finally, we formulate a platform-optimized hyperparameter calibration strategy to enable efficient training of the state-of-the-art (SOTA) transformer detector, RF-DETR~\cite{RF-DETR}, within the strict 16GB GPU memory limit. Our optimized variants, RF-DETR-HR and RF-DETR-Grayworld, bridge cross-city distribution gaps and achieve significant mAP scores of 47.53 and 46.63, respectively. We also achieve SOTA results and secure 1st place on the AIC2026-Track 6 leaderboard \cite{AIC2026Leaderboard}.

Our core technical contributions are summarized as follows:
\begin{itemize}
  \item We introduce a multi-dataset pre-training strategy with a class-agnostic objectness distillation pass to efficiently learn robust vehicle geometry.
  \item We present a domain-resilient augmentation pillar using a novel Grayworld transformation that forces global self-attention heads to remove transient chromatic shortcuts and favor robust shape priors.
  \item We construct two improved methods: RF-DETR-HR and RF-DETR-Grayworld, which achieve \textbf{SOTA, 1st place performance (mAP 47.53) on the hidden evaluation benchmark}.
\end{itemize}

\begin{figure}[tb]
  \centering
  \includegraphics[width=\textwidth]{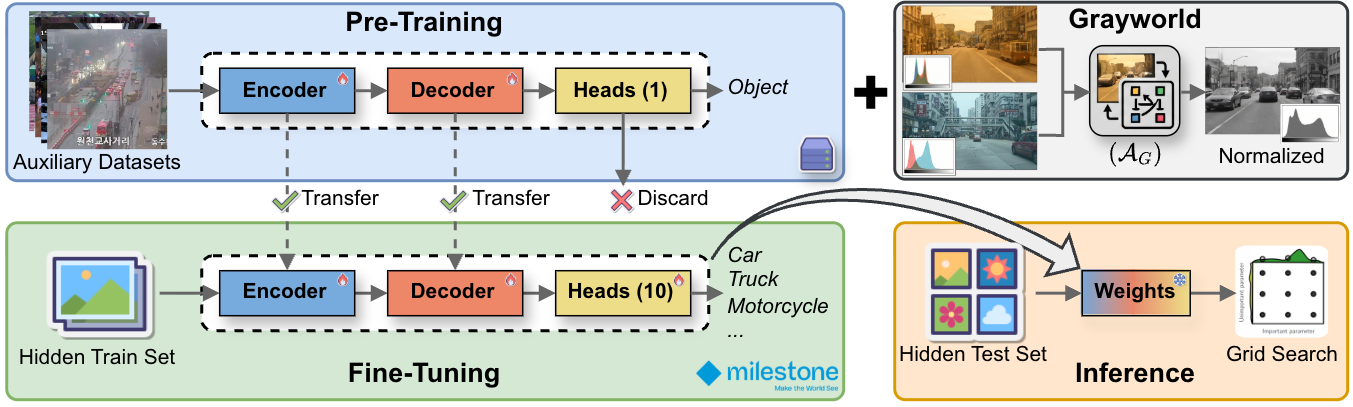}
  \caption{Overview of the proposed pre-training and Grayworld augmentation framework.}
  \label{fig:main}
\end{figure}

\section{Related Work}
\label{sec:related}

\subsection{Cross-Domain Object Detection}
\label{sec:object_detection}
Object detectors are primarily split into CNN-based (e.g., Faster R-CNN~\cite{FasterR-CNN}, YOLOs~\cite{YOLO, YOLO9000, YOLOv5, YOLOv7, YOLO26}) and transformer-based methods (e.g., DETR~\cite{DETR}, Deformable-DETR~\cite{Deformable-DETR}, Co-DETR~\cite{Co-DETR}, RF-DETR~\cite{RF-DETR}). While CNNs overfit to local source-domain features like lighting and asphalt style, transformers struggle with severe performance degradation under geographic domain shifts due to style-sensitive positional queries and attention maps. To bridge the gap between source and target distributions, existing literature relies on three main approaches: Unsupervised Domain Adaptation (UDA), Fully Test-Time Adaptation (F-TTA), and Vision-Language Model (VLM) prompting. UDA strategies~\cite{Tarvainen2017, Chen2018, Le2024} use adversarial learning, feature statistic alignment, or student-teacher pseudo-labeling to align feature distributions. F-TTA methods~\cite{TentFT, Chang2019} adapt model parameters online during inference via entropy minimization or batch-normalization updates. VLM-based approaches~\cite{CLIP, GroundingDINO, Qwen-VL, Qwen3} use textual prompts to guide visual representations toward a conceptual target domain.

Despite their success in standard benchmarks, these techniques fail when deployed under \textbf{true zero-shot, completely blind, and hardware-constrained settings}. UDA frameworks require offline access to unlabeled target-domain images during training. This assumption breaks in containerized or blind evaluation pipelines where target images remain completely hidden during the entire process. F-TTA methods execute backpropagation and gradient updates during runtime, which introduces severe constraints on resource-limited hardware and risks total model collapse if test-time pseudo-labels become noisy. Finally, VLM methods depend on prior knowledge of target conditions, which is unavailable in completely blind deployments. These disadvantages demonstrate the necessity of a purely source-side generalization strategy that enforces color-invariant, domain-agnostic feature representations during pre-training without requiring target data, test-time gradient updates, or prior domain assumptions.

\subsection{Traffic Surveillance Datasets}
\label{sec:datasets}
We describe the datasets used in this study and summarize them in Table \ref{tab:datasets}:

\begin{table}[t]
  \caption{Comparison of traffic surveillance datasets.}
  \label{tab:datasets}
  \centering
  \scriptsize
  \begin{tabular}{@{}llrrrrl@{}}
    \toprule
    \textbf{Dataset} & \textbf{Viewpoint} & \textbf{Images} & \textbf{Labels} & \textbf{Classes} & \textbf{Resolution} & \textbf{Characteristics} \\
    \midrule
    Hafnia~\cite{Hafnia-Dataset} & Coarse$\rightarrow$Fine & 27.8K & 151K  & 10 & 720--3648p  & Cross-city domain shift \\
    TSBOW~\cite{TSBOW}           & Coarse$\rightarrow$Fine & 48K   & 1.1M  & 8  & 720p       & Weather, time-of-day \\
    TrafficCAM~\cite{TrafficCAM} & Medium                  & 4.3K  & 84.2K & 10 & 288--1080p  & Complex traffic flow \\
    FishEye8K~\cite{FishEye8K}   & Medium                  & 8K    & 157K  & 5  & 1080--1280p & Distortion, omnidirection \\
    VisDrone~\cite{VisDrone}     & Coarse                  & 8.6K  & 2.6M  & 10 & 2160p      & Small objects \\
    MOT20~\cite{MOT20}           & Medium, Fine            & 13K   & 2.2M  & 1  & 1080p      & Pedestrian \\
    \bottomrule
  \end{tabular}
\end{table}
\begin{figure}[t]
  \centering
  \includegraphics[width=\textwidth]{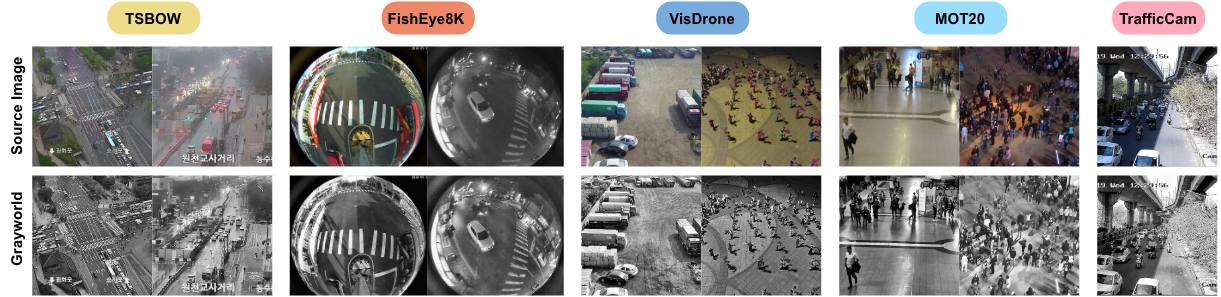}
  \caption{Visual comparisons of raw pre-training dataset images alongside their corresponding Grayworld-augmented variants. By neutralizing sensor-specific white-balance variances and volatile RGB distributions, the Grayworld transformation effectively bridges color and style discrepancies across disparate source domains.}
  \label{fig:augmentations2}
\end{figure}

\vspace{0.25em}
\noindent\textbf{Hafnia Dataset \cite{Hafnia-Dataset}} comprises 27.8K images (13K training and 15K testing) from a large-scale traffic dataset with diverse viewpoints, roadway types, camera perspectives, and imaging conditions. The test set contains images from the training city and a new target domain. The dataset contains 10 fine-grained object classes across two main categories: Vehicle and Person.

\vspace{0.25em}
\noindent\textbf{TSBOW~\cite{TSBOW}} is a diverse urban dataset featuring 48K manually labeled images, with an average of 122 objects per frame. It includes various weather conditions (sunny, rainy, snowy, and lighting) and scenarios (straight roads and intersections). This dataset provides models with useful prior knowledge and is well-suited to robust detection applications.

\vspace{0.25em}
\noindent\textbf{TrafficCAM~\cite{TrafficCAM}} is a collection depicting complex traffic flow in urban India. It contains top-down and side-frontal views at resolutions from 240p to FHD. The dataset includes over 4.3K images extracted from more than 2K videos under environmental conditions such as fog and nighttime.

\vspace{0.25em}
\noindent\textbf{FishEye8K~\cite{FishEye8K}} is a benchmark providing omnidirectional coverage using fisheye cameras. This dataset contains 8K images and 157K annotations under four illumination conditions (sunrise, sunny, sunset, and night) in Taiwan. The dataset has been widely used in previous AIC challenges~\cite{AIC2024, AIC2025, Pham2025}. Thus, it serves as an ideal training dataset to equip our proposed framework to detect objects under extreme lighting changes and diverse camera perspectives.

\vspace{0.25em}
\noindent\textbf{VisDrone~\cite{VisDrone}} consists of around 8.6K images at 4K resolution. This dataset is commonly used to improve detection and tracking models for small and medium objects captured by drone-mounted cameras~\cite{VisDrone-MOT2021}. It features highly crowded scenes with significant variations in object scale, occlusion, and perspective.

\vspace{0.25em}
\noindent\textbf{MOT20~\cite{MOT20}} is a benchmark for multi-person tracking in crowded scenes. This dataset contains 8 sequences with over 13K images, and the mean crowd density is 246 pedestrians per frame. Leveraging this data trains our framework to better distinguish overlapping person objects in dense crowds.

\section{Preliminary: Hafnia Training-aaS Platform}
\label{sec:hafnia}
Unlike conventional benchmark evaluations that test models locally on open-source data, AIC2026-Track 6 \cite{AIC2026} uses a privacy-preserving Milestone Systems' Hafnia TaaS platform~\cite{Hafnia}. It imposes the following constraints:
\begin{itemize}
  \item \textbf{No Data Access:} The training and benchmarking datasets are completely hidden. Algorithms must be packaged as standalone deployment units (i.e., Docker containers), and the data can only be accessed through anonymized training or inference jobs without debugging capability.
  \item \textbf{Blind Evaluation:} The benchmarking process is conducted in two steps. First, the Hafnia platform~\cite{Hafnia} performs inference on the testing images and generates prediction files. Then, these files are submitted to the AIC2026 evaluation server~\cite{AIC2026Leaderboard} to be evaluated against the hidden ground-truth.
  \item \textbf{Hardware Constraints:} Training and inference jobs are forced to run within deterministic resource, typically limited to x1--x4 NVIDIA Tesla T4 GPU (16GB VRAM) with a certain amount of training credits.
\end{itemize}
This TaaS platform changes the workflow from local open-loop training to a blind end-to-end training-inference pipeline. The hardware constraints also introduce a trade-off: VLM distillation~\cite{CLIP, GroundingDINO, Qwen-VL, Qwen3} becomes a computational bottleneck. Consequently, maximizing the mean Average Precision (mAP) requires localized feature calibration and a memory-efficient training pipeline rather than expanding backbones~\cite{YOLO26, DETR, Deformable-DETR, Co-DETR, RF-DETR} or complex F-TTA~\cite{TentFT, Chang2019}. Our proposed pipeline leverages this constraint as a design feature, optimizing training and inference pipelines to achieve maximum performance.

\section{Methodology}
\label{sec:method}

\subsection{Problem Formulation}
\label{sec:formulation}
We view the domain-robust object detection pipeline as a constrained optimization problem targeting generalization under strict geographic domain shift. Specifically, we systematically investigate the performance impact of different training and inference factors. We also need to consider memory consumption in practice. Let $\mathcal{F}$ denote the detection network parameterized by weights $\theta$. The objective is to maximize the mean Average Precision ($\text{mAP}$) over an unseen target domain distribution $\mathcal{D}_{T}$ without accessing its visual features or label spaces during both training and inference. Hence, we formulate this multi-stage optimization problem as:
\begin{equation}
  \begin{aligned}
    \max_{\mathcal{D}_S, \mathcal{A}, E, R_T, R_I, \tau_c} & \quad \text{mAP}
    (\mathcal{F} (\mathcal{D}_S, \mathcal{D}_T; \mathcal{A}, E, B, R_T, R_I, \tau_c )) \\
    \text{s.t.} & \quad \text{Memory}( \mathcal{F} (B, R_T, R_I)) \leq M_{\text{GPU}}
  \end{aligned}
  \label{eq:formulation}
\end{equation}
where:
\begin{itemize}
  \item $\mathcal{D}_S \in \{\mathcal{D}_1, \mathcal{D}_2, ... \}$ is our custom pre-training source dataset.
  \item $\mathcal{A}$ is the data augmentation strategy.
  \item $E$ is the number of training epochs.
  \item $B$ is the batch size during training.
  \item $R_T$ and $R_I$ are the training and inference resolutions, respectively.
  \item $\tau_c$ is the confidence threshold during inference.
\end{itemize}
It is obvious that either a larger model or a higher resolution increase both mAP and memory consumption. In practice, to fully utilize memory, we can attempt to calibrate $\mathcal{F}$, $E$, $B$, $R_T$, and $R_I$ to make the total memory consumption close to the maximum GPU memory ($M_{\text{GPU}}=\text{16GB}$). In the following sections, we focus on $\mathcal{D}_S$, $\mathcal{A}$, and the grid search approach.

\subsection{Multi-Dataset Pre-Training ($\mathcal{D}_S$)}
\label{sec:pretraining}

\begin{figure}[t]
  \centering
  \includegraphics[width=\textwidth]{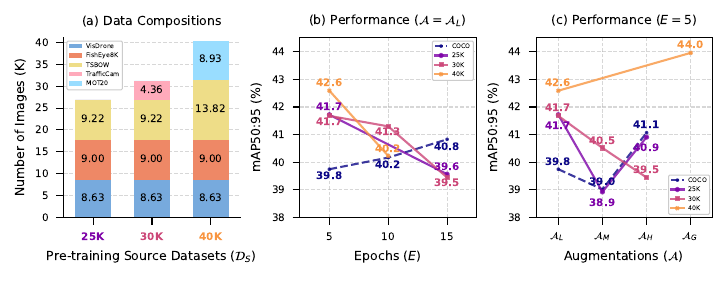}
  \caption{Compositions and performance comparison of auxiliary pre-training datasets ($\mathcal{D}_S$). Performance is evaluated on a 50\% subset of the hidden test set under identical inference configurations.}
  \label{fig:pretraining}
\end{figure}

Training detectors from scratch is time-consuming. Standard workflows rely on transfer learning from a large-scale distribution. However, generic distributions (e.g., COCO~\cite{COCO} or Objects365~\cite{Objects365}) have broader taxonomies than the specific, fine-grained categories in Table~\ref{tab:datasets}. Even when using traffic-specific datasets (in Section~\ref{sec:datasets}) for better initialization, class definitions still conflict across distributions. For example, a vehicle labeled as "Car" or "Van" in one dataset might be merged into a generic "Vehicle" class. To bridge this taxonomic gap, we discuss two main approaches: \textbf{pseudo-label} and \textbf{objectness distillation}.

\vspace{0.25em}
\noindent\textbf{Pseudo-label/Relabeling} aims to generate annotations for external data aligned with the target domain distribution. This process can be done via semi-supervised labeling~\cite{Lee2013, CVAT, LabelStudio} or VLM-based protocols~\cite{Qwen-VL, Qwen3, YOLO-World, Pham2024}. Specifically, we explored the zero-shot grounding and localization capabilities of Qwen3-VL~\cite{Qwen3}, denoted as $\mathcal{M}_{\text{VLM}}$, for processing unannotated images. By feeding images through $\mathcal{M}_{\text{VLM}}$ with structured prompt matrices, we dynamically generated high-confidence bounding box coordinates and fine-grained category logits. This automated relabeling strategy acts as a domain-specific filter, mapping open-vocabulary semantics onto the target category space to isolate relevant object features from out-of-distribution background noise. Ultimately, this technique harmonizes heterogeneous datasets into a synchronized label space. \textit{However, the generated pseudo-labels often require extensive manual refinement to reach the precision needed for fine-grained traffic surveillance. This makes the approach computationally feasible but labor-intensive.}

\vspace{0.25em}
\noindent\textbf{Objectness Distillation} aims to maximize the detector's capacity to identify fine-grained traffic primitives under significant distribution shifts. This method separates the learning of spatial geometry from category-specific classification. A class-agnostic structural distillation pass is implemented by collapsing all semantic labels from auxiliary datasets into a single binary objectness identifier (i.e., $\text{Class 0: object}$):
\begin{equation}
  \mathcal{Y}_{\text{agnostic}} = \{y_i \to 0 \mid \forall y_i \in \mathcal{Y}_{\text{source}}\}
\end{equation}
This formulation redefines the detection task as a binary foreground object localization problem. Applying this reduction over a warm-up schedule (3 to 5 epochs) forces the backbone encoder and transformer cross-attention layers to prioritize distilling high-fidelity object geometries through joint $L_1$ and GIoU loss minimization. This strategy prevents the classification head from overfitting early to the source city's specific vehicle distribution. By anchoring the model's queries on universal structural priors, we preserve the architectural plasticity required to rapidly adapt to the fine-grained 10-class taxonomy during the subsequent main training phases on the Hafnia platform.

To analyze the scaling behavior of our framework, we construct three pre-training datasets $\mathcal{D}_S \in \{\text{25K}, \text{30K}, \text{40K} \}$, aggregated from the heterogeneous traffic surveillance distributions detailed in Section~\ref{sec:datasets}. The exact composition and source breakdown for each subset are shown in Fig.~\ref{fig:pretraining}(a). We then conduct an empirical evaluation to assess the downstream generalization impact of each pre-training scale via a grid-search exploration in Section~\ref{sec:hyperparameter}. Based on observations from Figs.~\ref{fig:pretraining}(b, c), we use $\mathcal{D}_S = \text{40K}$ for the rest of the experiments in this paper.

\begin{figure}[t]
  \centering
  \includegraphics[width=\textwidth]{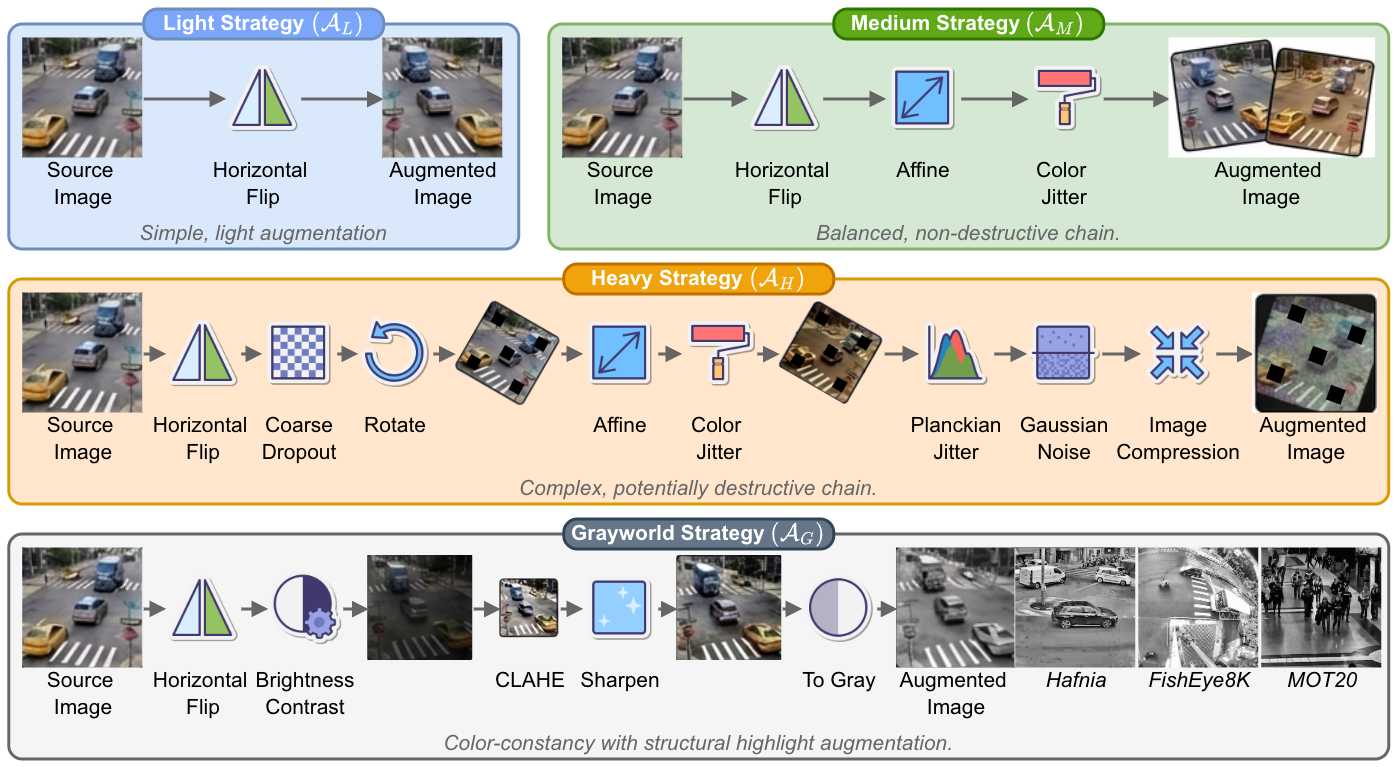}
  \caption{Illustration of augmentation strategies structured around order-specific sequential operations.}
  \label{fig:augmentations1}
\end{figure}

\subsection{Data Augmentation ($\mathcal{A}$)}
\label{sec:augmentation}
Under geographic domain shift, transformer-based architectures \cite{RF-DETR} often exhibit severe style memorization \cite{StyleProto}, where the encoder overfits to source-specific chromatic profiles, illumination artifacts, and sensor-level textures. To decouple vehicle geometry from environmental styles, we implement several standard augmentation strategies ($\mathcal{A}_L$, $\mathcal{A}_M$, $\mathcal{A}_H$) and propose a novel \textbf{Grayworld ($\mathcal{A}_G$)} approach. We follow Albumentations guideline \cite{Albumentations} to build augmentation pipelines. The detailed steps of each pipeline are listed in Fig.~\ref{fig:augmentations1}. Examples of augmented images are shown in Fig.~\ref{fig:augmentations2}.

\vspace{0.25em}
\noindent\textbf{Spatial-Geometric Augmentations ($\mathcal{A}_L$, $\mathcal{A}_M$, $\mathcal{A}_H$)} handle diverse camera viewpoints and scale distributions by applying randomized affine transformations such as rotation, horizontal flipping, and random cropping. These augmentations ensure the model is robust to layout variations typical of cross-city deployment.

\vspace{0.25em}
\noindent\textbf{Grayworld Augmentation ($\mathcal{A}_G$):} Geographic domain shift in TSS is driven by photometric variance across heterogeneous camera sensors (Fig.~\ref{fig:augmentations2}). Local illumination conditions, such as the cool-toned overcast light or the yellow chromaticity of nighttime sodium-vapor lamps, drastically perturb the RGB distributions of target objects. Transformer architectures \cite{RF-DETR} often lack intrinsic color constancy and are susceptible to shortcut learning. Instead of distilling domain-invariant geometric primitives, the global self-attention heads may overfit to source-specific chromatic footprints, like asphalt reflectance or camera white-balance biases. When deployed in an unseen target city, this learned dependency causes performance degradation as the target's chromatic profile diverges.

To address this problem, we draw inspiration from the robustness of grayscale thermal imaging \cite{Ahmar2025, TPMOT, SAGA, Tran2026}, where object detection relies exclusively on structural intensity and spatial gradients, thereby bypassing the volatility of chromatic signatures. This principle is mathematically aligned with the Gray-World Assumption \cite{Buchsbaum1980}, which states that the average reflectance of a complex scene is achromatic. Motivated by these concepts, we propose chromatic decoupling via grayscale intensity mapping and contrast enhancement. Since vehicle identifiers (e.g., chassis contours and wheel arches) are primarily geometric, they are preserved within high-frequency spatial gradients rather than color channels. Consequently, we collapse the RGB space into a single-channel intensity map throughout all training and inference phases. To compensate for the loss of color-based cues, we apply Contrast Limited Adaptive Histogram Equalization (CLAHE)~\cite{CLAHE} and edge sharpening, effectively forcing the network to prioritize robust shape-based semantics over transient, domain-specific color styles. Note that the Grayworld transformation must be applied consistently across pre-training, fine-tuning, and inference routines. Maintaining this strict grayscale consistency from initial weight anchoring to final blind deployment stabilizes the latent representation space, preventing feature drift and consistently increasing mAP on hidden benchmark streams, as verified in Section~\ref{sec:benchmark}.

\subsection{Hyperparameter Calibration}
\label{sec:hyperparameter}

\begin{figure}[t]
  \centering
  \includegraphics[width=\textwidth]{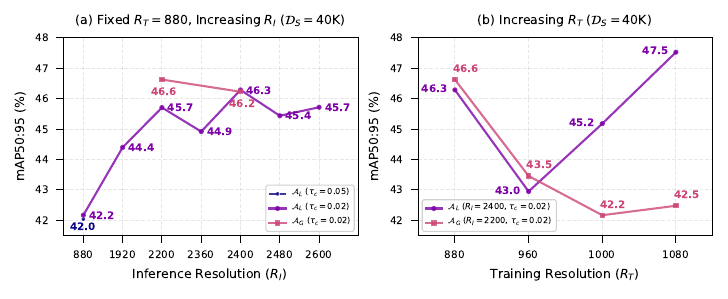}
  \caption{Impact of training resolution ($R_T$) and inference resolution ($R_I$) calibration on object detection performance. Performance is evaluated on the full hidden test set.}
  \label{fig:hyperparameters1}
\end{figure}

\begin{figure}[t]
  \centering
  \includegraphics[height=5.5cm]{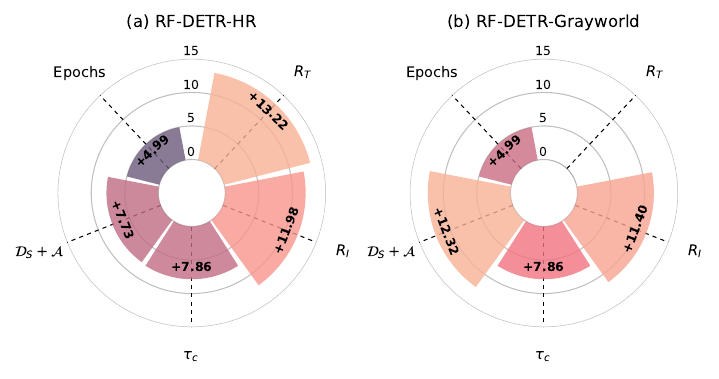}
  \caption{Incremental mAP performance contributions of each hyperparameter for \textbf{RF-DETR-HR} and \textbf{RF-DETR-Grayworld} relative to the COCO pre-trained RF-DETR-2XLarge baseline. Axes are arranged in counter-clockwise order.}
  \label{fig:improve_methods}
\end{figure}

Solving the global optimization objective in Eq.~\ref{eq:formulation} is challenging due to the high-dimensional parameter space and complex cross-domain dynamics. To find effective training and inference configurations under platform constraints, we conduct a systematic grid-search exploration. We evaluate the sensitivity and cumulative impact of seven core factors: model size ($\mathcal{F}$), pre-training data ($\mathcal{D}_S$), augmentation strategy ($\mathcal{A}$), training epochs ($E$), training resolution ($R_T$), inference resolution ($R_I$), and confidence threshold ($\tau_c$). The complete experimental matrix and empirical trends are summarized in Table~\ref{tab:experiment} and Figs.~\ref{fig:hyperparameters1}--\ref{fig:improve_methods}. The specific factor configurations and key findings are detailed as follows.

\vspace{0.25em}
\noindent \textbf{Model Size ($\mathcal{F}$):} Scaling the detector's backbone capacity is the most direct way to improve raw mAP under severe domain shift. However, larger variants consume a large part of the platform's 16GB VRAM budget. To find the optimal compute allocation, we stress-tested all RF-DETR~\cite{RF-DETR} variants within the Hafnia platform. We find that optimizing the largest model variant (RF-DETR-2XLarge) is feasible by limiting the physical batch size per GPU to $B=2$. To reduce high gradient variance and match the optimization trajectory of a larger batch, we combine this with an 8-step gradient accumulation strategy, keeping an effective batch size of 16.

\vspace{0.25em}
\noindent \textbf{Pre-training ($\mathcal{D}_S$) and Augmentation ($\mathcal{A}$):} We conduct a multi-factorial grid search between auxiliary pre-training $\mathcal{D}_S \in \{\text{25K}, \text{30K}, \text{40K}\}$ and augmentation strategy $\mathcal{A} \in \{\mathcal{A}_L, \mathcal{A}_M, \mathcal{A}_H, \mathcal{A}_G\}$. Our observations show that increasing pre-training volume provides richer spatial priors and diverse vehicle viewpoint geometries, yielding steady mAP improvements on unseen target cities (Fig.~\ref{fig:pretraining}). Crucially, we identify a distinct coupling: aggressive augmentation schemes ($\mathcal{A}_H$) are effective mainly when transferring from generic COCO initializations to prevent early feature over-consolidation. Conversely, when the model is pre-trained on our 40K traffic dataset, light augmentations ($\mathcal{A}_L$) and Grayworld ($\mathcal{A}_G$) achieve optimal convergence, preserving distilled vehicle structural features while neutralizing domain-specific style drift.

\vspace{0.25em}
\noindent \textbf{Training Epochs ($E$):} In ZCOD deployment, the training schedule controls the trade-off between domain feature adaptation and source-domain style memorization. Our grid search shows that a compact schedule of $E = 5$ epochs yields optimal cross-city mAP performance. Extending fine-tuning beyond this (e.g., $E \in \{10, 15\}$) causes a noticeable decline in target-domain accuracy (Fig.~\ref{fig:pretraining}(b)).

\vspace{0.25em}
\noindent \textbf{Training and Inference Resolutions ($R_T, R_I$):} Image resolution directly dictates feature map granularity and positional query alignment in Vision Transformers. We evaluate training resolutions $R_T \in \{880, 960, 1000, 1080\}$ and inference resolutions $R_I \in \{880, 1920, 2200, 2360, 2400, 2480, 2600\}$. Our empirical grid search reveals three key insights about resolution dynamics:
\begin{itemize}
  \item \textbf{Inference Scale Saturation:} mAP increases with inference resolution $R_I$ as higher pixel densities restore fine-grained spatial cues for small vehicles. However, performance reaches a saturation point, peaking at $2400 \times 2400$ for the light augmentation ($\mathcal{A}_L$) and $2200 \times 2200$ for the Grayworld augmentation ($\mathcal{A}_G$) before over-smoothing artifacts and logit noise reduce precision.
  \item \textbf{Augmentation-Scale Coupling:} Increasing the training resolution $R_T$ to $1080 \times 1080$ mainly benefits $\mathcal{A}_L$ by expanding spatial feature resolution. In contrast, models trained with $\mathcal{A}_G$ show intrinsic scale-invariant shape expression and maintain strong generalization without high training resolutions.
  \item \textbf{Platform Computation:} Scaling training resolution beyond the $880 \times 880$ baseline causes significant memory overhead, forcing batch size reduction to $B=1$ to avoid out of memory within the 16GB VRAM limit. Increasing inference scale to $R_I = 2600$ remains viable on the Hafnia deployment container by optimizing the model's forward execution graph via TorchScript Just-In-Time (JIT) tracing with dynamic precision casting.
\end{itemize}

\noindent \textbf{Confidence Threshold ($\tau_c$):} The confidence threshold controls the trade-off between false-positive filtering and target recall during test-time evaluation. In ZCOD deployment, models often show under-confident prediction logits due to out-of-distribution. Our grid search shows that lowering the post-processing threshold ($\tau_c = 0.02$) consistently increases overall mAP. By relaxing $\tau_c$, the detector keeps valid, low-confidence object queries such as heavily occluded, small, or distant vehicles that standard thresholds ($\tau_c \ge 0.25$) would prune aggressively. Since mAP evaluates the entire area under the Precision-Recall curve across IoU thresholds, recovering these true positive detections at lower confidence levels significantly boosts overall recall without large precision penalties.

\subsection{Improved Methods}
\label{sec:improved}

Building upon the empirical findings and hyperparameter dynamics analyzed in Sections~\ref{sec:pretraining}--\ref{sec:hyperparameter}, we construct two model configurations tailored for ZCOD:
\begin{itemize}
  \item \textbf{RF-DETR-HR:} Integrates domain-aligned 40K pre-training with light data augmentation ($\mathcal{A}_L$), high-resolution training ($R_T = 1080 \times 1080$), and test-time inference scaling ($R_I = 2400 \times 2400$), maximizing spatial feature density to resolve small and distant traffic targets.
  \item \textbf{RF-DETR-Grayworld:} Combines domain-aligned 40K pre-training and Grayworld chromatic transformation ($\mathcal{A}_G$) alongside test-time inference scaling ($R_I = 2200 \times 2200$), forcing the model to prioritize domain-invariant structural geometry over volatile, location-specific color statistics.
\end{itemize}
The relative performance contributions and empirical ablation trajectories of each component are visually summarized in Fig.~\ref{fig:improve_methods}.

\section{Experimental Results}
\label{sec:experiment}

\begin{table}[t]
  \begin{threeparttable}
    \caption{Experimental results on the full test set of AIC2026-Track~6~\cite{AIC2026Leaderboard}. The best and the second-best results are in \textbf{bold} and \underline{underline}, respectively.}
    \label{tab:experiment}
    \centering
    \scriptsize
    \begin{tabular*}{\textwidth}{@{\extracolsep{\fill}} ll|cccccc|ccc}
      \toprule
      \textbf{\#} & \textbf{Model} ($\mathcal{F}$) & $\mathcal{D}_S$ & $E$ & $\mathcal{A}$ & $R_T$ & $R_I$ & $\tau_c$ & \textbf{mAP} & \textbf{mAP50} & \textbf{mAP75} \\
      \midrule
      0  & RF-DETR-Nano (Baseline) & COCO & 8 & - & 384 & 384 & 0.05 & 23.24 & 33.77 & 24.13 \\
      \midrule
      1  & RF-DETR-Nano            & COCO & 1 & - & 384 & 384 & 0.05 & 26.72 & 38.10 & 27.11 \\
      2  & RF-DETR-Large           & COCO & 1 & - & 704 & 704 & 0.05 & 30.77 & 44.92 & 32.24 \\
      3  & RF-DETR-XLarge          & COCO & 1 & - & 704 & 704 & 0.05 & 31.06 & 43.39 & 33.00 \\
      4  & RF-DETR-2XLarge         & COCO & 1 & - & 880 & 880 & 0.05 & 34.31 & 46.99 & 36.61 \\
      \midrule
      5  & RF-DETR-2XLarge         & COCO & 5 & $\mathcal{A}_L$ & 880  & 880  & 0.05 & 39.30 & 53.18 & 41.31 \\
      6  & RF-DETR-2XLarge         & COCO & 5 & $\mathcal{A}_M$ & 880  & 880  & 0.05 & 38.43 & 52.88 & 40.47 \\
      7  & RF-DETR-2XLarge         & COCO & 5 & $\mathcal{A}_H$ & 880  & 880  & 0.05 & 40.37 & 56.11 & 43.04 \\
      \midrule
      8  & RF-DETR-2XLarge         & 25K  & 5 & $\mathcal{A}_L$ & 880  & 880  & 0.05 & 42.28 & 56.85 & 43.95 \\
      9  & RF-DETR-2XLarge         & 25K  & 5 & $\mathcal{A}_M$ & 880  & 880  & 0.05 & 40.50 & 55.03 & 43.28 \\
      10 & RF-DETR-2XLarge         & 25K  & 5 & $\mathcal{A}_H$ & 880  & 880  & 0.05 & 38.43 & 54.68 & 39.90 \\
      \midrule
      11 & RF-DETR-2XLarge         & 30K  & 5 & $\mathcal{A}_L$ & 880  & 880  & 0.05 & 41.18 & 55.26 & 43.18 \\
      12 & RF-DETR-2XLarge         & 40K  & 5 & $\mathcal{A}_L$ & 880  & 880  & 0.05 & 42.04 & 55.58 & 44.10 \\
      \midrule
      13 & RF-DETR-2XLarge         & 40K  & 5 & $\mathcal{A}_L$ & 880  & 880  & 0.02 & 42.17 & 55.80 & 44.23 \\
      14 & RF-DETR-2XLarge         & 40K  & 5 & $\mathcal{A}_L$ & 880  & 1920 & 0.02 & 44.40 & 57.71 & 47.65 \\
      15 & RF-DETR-2XLarge         & 40K  & 5 & $\mathcal{A}_L$ & 880  & 2200 & 0.02 & 45.71 & 58.86 & 48.86 \\
      16 & RF-DETR-2XLarge         & 40K  & 5 & $\mathcal{A}_L$ & 880  & 2400 & 0.02 & 46.29 & 59.51 & 49.84 \\
      17 & RF-DETR-2XLarge         & 40K  & 5 & $\mathcal{A}_L$ & 960  & 2400 & 0.02 & 42.95 & 57.41 & 45.71 \\
      18 & RF-DETR-2XLarge         & 40K  & 5 & $\mathcal{A}_L$ & 1000 & 2400 & 0.02 & 45.18 & 59.19 & 48.13 \\
      \textbf{19} & \textbf{RF-DETR-2XLarge} & \textbf{40K} & \textbf{5} & \textbf{$\mathcal{A}_L$} & \textbf{1080} & \textbf{2400} & \textbf{0.02} & \textbf{47.53} & \textbf{62.07} & \textbf{50.31} \\
      & (RF-DETR-HR) & & & & & & & & & \\
      \midrule
      \underline{20} & \underline{RF-DETR-2XLarge} & \underline{40K} & \underline{5} & \underline{$\mathcal{A}_G$} & \underline{880} & \underline{2200} & \underline{0.02} & \underline{46.63} & \underline{59.46} & \underline{49.96} \\
      & (RF-DETR-Grayworld) & & & & & & & & & \\
      21 & RF-DETR-2XLarge         & 40K  & 5 & $\mathcal{A}_G$ & 880  & 2400 & 0.02 & 46.23 & 59.39 & 49.23 \\
      22 & RF-DETR-2XLarge         & 40K  & 5 & $\mathcal{A}_G$ & 960  & 2200 & 0.02 & 43.46 & 57.01 & 47.14 \\
      23 & RF-DETR-2XLarge         & 40K  & 5 & $\mathcal{A}_G$ & 1000 & 2200 & 0.02 & 42.17 & 55.62 & 44.87 \\
      24 & RF-DETR-2XLarge         & 40K  & 5 & $\mathcal{A}_G$ & 1080 & 2200 & 0.02 & 42.48 & 55.89 & 45.12 \\
      \bottomrule
    \end{tabular*}
  \end{threeparttable}
\end{table}

\subsection{Implementation Details}
\label{sec:implementation}

\noindent\textbf{Local Pre-training} is conducted on an NVIDIA RTX A6000 GPU (48GB VRAM). For each pre-training dataset $\mathcal{D}_S \in \{\text{25K}, \text{30K}, \text{40K}\}$, we optimize two sets of weights: an unaugmented baseline and a Grayworld variant. Furthermore, whenever the training resolution $R_T$ is adjusted, we need to pre-train dedicated checkpoint weights to preserve scale-dependent positional embeddings. This warm-up phase runs for a brief 3--5 epoch schedule, after which the top-performing Exponential Moving Average (EMA) weights are selected as the initialization checkpoint for downstream fine-tuning. 

\vspace{0.25em}
\noindent\textbf{Online Fine-tuning} is performed on the hidden training split within the Hafnia Training-aaS environment~\cite{Hafnia} using an NVIDIA Tesla T4 GPU (16GB VRAM). We load the localization-anchored pre-trained weights and adapt the classification prediction heads to the full 10-class target. To ensure spatial feature adaptation while maintaining semantic stability, we use a differential learning rate strategy. We assign a base learning rate of $\text{lr} = 10^{-4}$ to the transformer decoder and classification heads, and a slightly higher rate of $\text{lr}_{\text{enc}} = 1.5 \times 10^{-4}$ to the transformer encoder. This asymmetrical setup prioritizes spatial context adaptation, which is essential for accommodating novel camera viewpoints and road layouts in unseen target cities. At the same time, it conservatively updates the classification heads to reduce catastrophic forgetting of pre-trained spatial priors. The complete set of fine-tuning hyperparameters is detailed in Section~\ref{sec:hyperparameter}.

\vspace{0.25em}
\noindent\textbf{Benchmarking} process is conducted in two steps. First, the containerized model runs forward-pass inference on the concealed target-city test split hosted on the Hafnia platform infrastructure~\cite{Hafnia}, producing standardized prediction files. Then, these prediction files are submitted to the official AIC2026 evaluation server~\cite{AIC2026Leaderboard} for hidden ground-truth benchmarking and leaderboard ranking. The primary evaluation is based on standard COCO-Eval \cite{COCO} metrics, with the final ranking determined by mAP50:95. 


\subsection{Benchmark Results}
\label{sec:benchmark}
In this section, we present our empirical benchmarking framework (results shown in Table~\ref{tab:experiment}), detailing the iterative experimental setup, calibration progression, and key methodological findings derived across each evaluation cycle:

\vspace{0.25em}
\noindent\textbf{Model Scale Exploration (\#1--4):} We evaluated all RF-DETR model variants by fine-tuning each architecture for one epoch using standard COCO pre-trained weights under default training settings. Our initial benchmark showed that the largest variant, RF-DETR-2XLarge, achieved the highest target-domain mAP ($34.31$) among the baselines. Furthermore, our ablation study (in Fig.~\ref{fig:pretraining}(b)) confirms that restricting fine-tuning to $E \le 5$ epochs significantly outperformed longer schedules ($26.72$ vs.\ $23.24$ mAP), effectively preventing transformer attention heads from memorizing source-city visual clutter and overfitting to domain-specific artifacts.

\vspace{0.25em}
\noindent\textbf{Pre-training and Augmentation (\#5--12):} We systematically evaluated the combinations of auxiliary pre-training datasets $\mathcal{D}_S$ and augmentation strategies $\mathcal{A}$. Our findings reveal a key paradigm: scaling the domain-aligned pre-training dataset to 40K instances yields a significantly higher cross-city mAP ($42.04$) than applying aggressive synthetic augmentations ($\mathcal{A}_H$) on smaller initializations ($40.37$). Heavy synthetic transformations risk corrupting feature geometries, whereas expanding real-world pre-training volume exposes the model to natural environmental, lighting, and viewpoint diversity. Thus, expanding the pre-training data acts as a powerful macro-level augmentation technique, providing realistic domain invariance while preserving structural fidelity.

\vspace{0.25em}
\noindent\textbf{Hyperparameter Calibration (\#13--19):} After identifying the optimal pre-training and augmentation strategy, we calibrated image resolutions ($R_T, R_I$) and post-processing thresholds ($\tau_c$). First, lowering the confidence threshold to $\tau_c = 0.02$ recovered under-confident target predictions caused by domain-shift logit suppression, yielding a slight mAP increase ($+0.13$). Second, increasing test-time inference resolution $R_I$ to $2400 \times 2400$ significantly boosted small-target recall ($46.29$ vs.\ $42.17$). Finally, increasing training resolution $R_T$ to $1080 \times 1080$ further expanded spatial feature map capacity for the $\mathcal{A}_L$ augmentation ($+1.24$ mAP), though this required reducing the physical batch size ($B=1$) to strictly fit the 16GB VRAM limit.

\vspace{0.25em}
\noindent\textbf{Grayworld Augmentation (\#20--24):} We also explored the impact of the Grayworld chromatic transformation ($\mathcal{A}_G$) applied during local pre-training and online fine-tuning. Our results show that by removing volatile RGB color distributions and neutralizing camera white-balance variances, Grayworld augmentation prevents the Vision Transformer from exploiting transient photometric shortcuts. Specifically, at the same training resolution of $R_T = 880$, the Grayworld variant outperforms light augmentation baselines ($46.63$ vs.\ $45.71$ mAP), confirming its robustness against unpredictable lighting and sensor style shifts.

\begin{table}[t]
  \begin{threeparttable}
    \caption{Official public leaderboard for AIC2026-Track 6~\cite{AIC2026Leaderboard}. Our team, SKKU-AL-T1 (Team ID 34), secures 1st place with an absolute mAP margin of $+4.72$ over the second-place entry.}
    \label{tab:leaderboard}
    \centering
    \scriptsize
    \begin{tabular*}{\textwidth}{@{\extracolsep{\fill}} ccll|ccc}
      \toprule
      \textbf{Rank} & \textbf{Team ID} & \textbf{Team Name} & \textbf{Model} & \textbf{mAP} & \textbf{mAP50} & \textbf{mAP75} \\
      \midrule
      \textbf{1} & \textbf{34} & \textbf{SKKU-AL-T1} & \textbf{RF-DETR-HR} & \textbf{47.53} & \textbf{62.07} & \textbf{50.31} \\
         &     &              & \underline{RF-DETR-Grayworld} & \underline{46.63} & \underline{59.46} & \underline{49.96} \\
      \midrule
      2  & 256 & BIT-ODL      & detr                       & 42.81 & 55.39 & 45.68 \\
      3  & 152 & Chisinau     & RF-DETR Large              & 41.76 & 57.27 & 45.21 \\
      4  & 261 & BK2TheFuture & RFDETR2XLarge              & 41.69 & 52.89 & 44.07 \\
      5  & 162 & S2 Detection & RF-DETR                    & 41.14 & 54.47 & 44.14 \\
      6  & 58  & VisionOps    & detr ensemble              & 40.22 & 53.33 & 42.59 \\
      7  & 273 & Team IPCV    & DEIMv2X, Dinov3S+          & 38.82 & 50.21 & 40.68 \\
      8  & 101 & bfc          & detr                       & 37.41 & 50.97 & 39.89 \\
      9  & 247 & NextITS      & RF-DETR Large, square 1120 & 36.54 & 48.79 & 38.57 \\
      10 & 188 & Team United  & RF-DETR Large              & 35.29 & 46.55 & 36.56 \\
      \bottomrule
    \end{tabular*}
  \end{threeparttable}
\end{table}

\subsection{Comparison \& Final Ranking}
\label{sec:ranking}
The final rankings of AIC2026-Track 6 \cite{AIC2026Leaderboard} are summarized in Table~\ref{tab:leaderboard}. Our solutions, RF-DETR-HR and RF-DETR-Grayworld, rank 1st ($47.53$ mAP) and 2nd nominally ($46.63$ mAP) according to the reported results. Furthermore, our methods outperform the subsequent competitor entry by substantial absolute margins ($+4.7$ and $+3.8$ mAP). 

\section{Conclusion}
\label{sec:conclusion}
In this work, we addressed zero-shot cross-city object detection under strict hardware and privacy constraints on the Milestone System' Hafnia platform. To tackle geographic domain shifts without changing complex Vision Transformer architectures, we introduced a modular framework based on two pillars: a class-agnostic 40K pre-training strategy that separates vehicle structural geometry from changing taxonomies, and a Grayworld chromatic transformation that removes sensor-dependent color shortcuts. Combined with platform-optimized techniques such as differential learning rates, gradient accumulation, and JIT tracing, our optimized models (RF-DETR-HR and RF-DETR-Grayworld) operate efficiently within a 16GB VRAM limit. These innovations yielded substantial mAP gains over standard baselines and secured the \textbf{1st place on the official AI City Challenge (AIC2026) Track 6 evaluation leaderboard}.

\section*{Acknowledgements}
This work was supported by Institute of Information \& Communications Technology Planning \& Evaluation (IITP) grant funded by the Korea government (MSIT) (No.2021-0-01364, An intelligent system for 24/7 real-time traffic surveillance on edge devices)

%
\bibliographystyle{splncs04}
\bibliography{main}

@String(CVPR  = {IEEE Conf. Comput. Vis. Pattern Recog.})

@String(ICCV  = {Int. Conf. Comput. Vis.})

@String(ECCV  = {Eur. Conf. Comput. Vis.})

@String(ICML  = {Int. Conf. Mach. Learn.})

@String(ICLR  = {Int. Conf. Learn. Represent.})

@String(AAAI  = {AAAI})

@String(CVPR  = {CVPR})

@String(ICCV  = {ICCV})

@String(ECCV  = {ECCV})

@String(ICML  = {ICML})

@String(ICLR  = {ICLR})

@inproceedings{Ahmar2025,
    author    = {Ahmar, Wassim A. El and Kolhatkar, Dhanvin and Nowruzi, Farzan Erlik and Laganiere, Robert},
    title     = {{Enhancing Thermal MOT: A Novel Box Association Method Leveraging Thermal Identity and Motion Similarity}},
    booktitle = {European Conference on Computer Vision (ECCV) Workshops},
    year      = {2024},
    pages     = {103--120},
}

@InProceedings{AIC2024,
    author    = {Shuo Wang and David C. Anastasiu and Zheng Tang and Ming-Ching Chang and others},
    title     = {{The 8th AI City Challenge}},
    booktitle = {The IEEE Conference on Computer Vision and Pattern Recognition (CVPR) Workshops},
    year      = {2024},
}

@InProceedings{AIC2025,
    author    = {Tang, Zheng and Wang, Shuo and Anastasiu, David C. and Chang, Ming-Ching and others},
    title     = {{The 9th AI City Challenge}},
    booktitle = {IEEE/CVF International Conference on Computer Vision (ICCV) Workshops},
    year      = {2025},
    pages     = {5526-5535},
}

@inproceedings{AIC2026,
    author    = {Tang, Zheng and Wang, Shuo and Anastasiu, David C. and Chang, Ming-Ching and others},
    title     = {The 10th {AI City Challenge}},
    booktitle = {European Conference on Computer Vision (ECCV) Workshops},
    year      = {2026},
    location  = {Malm{"o}, Sweden},
}

@misc{AIC2026Leaderboard,
    title = {{AI City Challenge Evaluation System}},
    year  = {2026},
    url   = {https://eval.aicitychallenge.org/aicity2026/submission/leaderboard},
}

@article{Albumentations,
    author         = {Buslaev, Alexander and Iglovikov, Vladimir I. and Khvedchenya, Eugene and others},
    title          = {{Albumentations: Fast and Flexible Image Augmentations}},
    journal        = {Information},
    volume         = {11},
    year           = {2020},
    number         = {2},
    article-number = {125},
    issn           = {2078-2489},
}

@article{Buchsbaum1980,
    title     = {{A Spatial Processor Model for Object Colour Perception}},
    author    = {Buchsbaum, Gershon},
    journal   = {Journal of the Franklin Institute},
    volume    = {310},
    number    = {1},
    pages     = {1--26},
    year      = {1980},
}

@inproceedings{Chang2019,
  author    = {Chang, Woong-Gi and You, Takuhiro and Seo, Seonguk and others},
  title     = {{Domain-Specific Batch Normalization for Unsupervised Domain Adaptation}},
  booktitle = {IEEE/CVF Conference on Computer Vision and Pattern Recognition (CVPR)},
  year      = {2019},
}

@inproceedings{Chen2018,
  author    = {Chen, Yuhua and Li, Wen and Sakaridis, Christos and others},
  title     = {{Domain Adaptive Faster R-CNN for Object Detection in the Wild}},
  booktitle = {IEEE/CVF Conference on Computer Vision and Pattern Recognition (CVPR)},
  year      = {2018},
}

@incollection{CLAHE,
    title     = {{Contrast-Limited Adaptive Histogram Equalization}},
    author    = {Zuiderveld, Karel},
    booktitle = {Graphics gems IV},
    pages     = {474--485},
    year      = {1994},
    publisher = {Academic Press Professional, Inc.},
}

@InProceedings{CLIP,
  title     = {{Learning Transferable Visual Models From Natural Language Supervision}},
  author    = {Radford, Alec and others},
  booktitle = {38th International Conference on Machine Learning (ICML)},
  pages     = {8748--8763},
  year      = {2021},
}

@inproceedings{COCO,
    title     = {{Microsoft COCO: Common Objects in Context}},
    author    = {Lin, Tsung-Yi and Maire, Michael and Belongie, Serge and others},
    booktitle = {European Conference on Computer Vision (ECCV)},
    pages     = {740--755},
    year      = {2014},
}

@inproceedings{Co-DETR,
  title     = {{DETRs with Collaborative Hybrid Assignments Training}},
  author    = {Zong, Zhuofan and Song, Guanglu and Liu, Yuhang and others},
  booktitle = {IEEE/CVF International Conference on Computer Vision (ICCV)},
  pages     = {6748--6758},
  year      = {2023},
}

@software{CVAT,
    author    = {Sekachev, Boris and others},
    title     = {{Computer Vision Annotation Tool (CVAT)}},
    year      = 2020,
    publisher = {Zenodo},
    version   = {v1.1.0},
}

@inproceedings{Deformable-DETR,
    title     = {{Deformable DETR: Deformable Transformers for End-to-End Object Detection}},
    author    = {Xizhou Zhu and Weijie Su and Lewei Lu and others},
    booktitle = {International Conference on Learning Representations},
    year      = {2021},
}

@inproceedings{DETR,
  title     = {{DETRs Beat YOLOs on Real-time Object Detection}},
  author    = {Zhao, Yian and Lv, Wenyu and Xu, Shangliang and others},
  booktitle ={IEEE/CVF Conference on Computer Vision and Pattern Recognition (CVPR)},
  pages     = {12739--12748},
  year      = {2024},
}

@article{FasterR-CNN,
  title   = {{Faster R-CNN: Towards Real-Time Object Detection with Region Proposal Networks}},
  author  = {Ren, Shaoqing and He, Kaiming and Girshick, Ross and Sun, Jian},
  journal = {IEEE Transactions on Pattern Analysis and Machine Intelligence},
  volume  = {39},
  number  = {6},
  pages   = {1137--1149},
  year    = {2017},
}

@inproceedings{FishEye8K,
    author    = {Gochoo, Munkhjargal and Otgonbold, Munkh-Erdene and Ganbold, Erkhembayar and others},
    title     = {{FishEye8K: A Benchmark and Dataset for Fisheye Camera Object Detection}},
    booktitle = {IEEE/CVF Conference on Computer Vision and Pattern Recognition (CVPR) Workshops},
    year      = {2023},
    pages     = {5304--5312},
}

@inproceedings{GroundingDINO,
  title     = {{Grounding DINO: Marrying DINO with Grounded Pre-Training for Open-Set Object Detection}},
  author    = {Liu, Shilong and Zeng, Zhaoyang and Ren, Tianhe and others},
  booktitle = {European Conference on Computer Vision (ECCV)},
  year      = {2024},
}

@misc{Hafnia,
    author = {{Milestone Systems}},
    title  = {{Project Hafnia: A Game-Changer in AI Model Training}},
    year   = {2025},
}

@misc{Hafnia-Dataset,
    author  = {{Milestone Systems}},
    title   = {{Hafnia Dataset: ECCV Cross-City Object Detection Dataset}},
    note    = {Part of the Hafnia project},
    version = {0.0.1},
    year    = {2024},
}

@misc{LabelStudio, 
    author = {Tkachenko, M. and others}, 
    title  = {{Label Studio: Data Labeling Software}}, 
    year   = {2020}, 
}

@inproceedings{Le2024,
  title     = {{Dynamic Retraining-Updating Mean Teacher for Source-Free Object Detection}},
  author    = {Khanh, Trinh Le Ba and Nguyen, Huy-Hung and Pham, Long Hoang and others},
  booktitle = {European Conference on Computer Vision (ECCV)},
  pages     = {328--344},
  year      = {2024},
}

@inproceedings{Lee2013,
    title     = {{Pseudo-label: The Simple and Efficient Semi-supervised Learning Method for Deep Neural Networks}},
    author    = {Lee, Dong-Hyun},
    booktitle = {Workshop on challenges in representation learning, ICML},
    volume    = {3},
    number    = {2},
    pages     = {896},
    year      = {2013},
}

@article{MOT20,
    title   = {{MOT20: A Benchmark for Multi-object Tracking in Crowded Scenes}},
    author  = {Dendorfer, Patrick and Rezatofighi, Hamid and Milan, Anton and others},
    journal = {arXiv preprint arXiv:2003.09003},
    year    = {2020},
}

@inproceedings{Objects365,
    author    = {Shao, Shuai and Li, Zeming and Zhang, Tianyuan and others},
    booktitle = {IEEE/CVF International Conference on Computer Vision (ICCV)}, 
    title     = {{Objects365: A Large-Scale, High-Quality Dataset for Object Detection}}, 
    year      = {2019},
    pages     = {8429--8438},
  }

@InProceedings{Pham2024,
    author    = {Pham, Long Hoang and Ho, Quoc Pham-Nam and Tran, Duong Nguyen-Ngoc and others},
    title     = {{Improving Object Detection to Fisheye Cameras with Open-Vocabulary Pseudo-Label Approach}},
    booktitle = {IEEE/CVF Conference on Computer Vision and Pattern Recognition (CVPR) Workshops},
    year      = {2024},
    pages     = {7100-7109},
}

@inproceedings{Pham2025,
  title     = {{Data Augmentation Is All You Need For Robust Fisheye Object Detection}},
  author    = {Pham, Long Hoang and Ho, Quoc Pham-Nam and Vu, Duong Khac and others},
  booktitle = {Proceedings of the IEEE/CVF International Conference on Computer Vision (ICCV) Workshops},
  year      = {2025},
}

@article{Qwen-VL,
    title   = {{Qwen-VL: A Versatile Vision-Language Model for Understanding, Localization, Text Reading, and Beyond}},
    author  = {Bai, Jinze and Bai, Shuai and Yang, Shusheng and others},
    journal = {arXiv preprint arXiv:2308.12966},
    year    = {2023},
}

@misc{Qwen3,
    title         = {{Qwen3 Technical Report}}, 
    author        = {{Qwen Team}},
    year          = {2025},
    eprint        = {2505.09388},
    archivePrefix = {arXiv},
}

@inproceedings{RF-DETR,
    title     = {{RF-DETR: Real-Time Detection Transformer}},
    author    = {Robinson, Isaac and Robicheaux, Peter and Popov, Matvei and others},
    booktitle = {International Conference on Learning Representations (ICLR)},
    year      = {2026},
}

@inproceedings{SAGA,
    title     = {{SAGA: Semantic-Aware Gray Color Augmentation for Visible-to-Thermal Domain Adaptation Across Multi-View Drone and Ground-Based Vision Systems}},
    author    = {Manjunath, D and Sikdar, Aniruddh and Gurunath, Prajwal and others},
    booktitle = {IEEE/CVF Conference on Computer Vision and Pattern Recognition (CVPR) Workshops},
    pages     = {4578--4588},
    year      = {2025},
}

@article{Saunier2018,
    title   = {{An Embedded Computer-Vision System for Multi-Object Detection in Traffic Surveillance}},
    author  = {Saunier, Nicolas and others},
    journal = {IEEE Transactions on Intelligent Transportation Systems},
    year    = {2018},
}

@inproceedings{StyleProto,
    title     = {{StyleProto: Style-Augmented Prototype Learning for Cross-Domain Few-Shot Object Detection}},
    author    = {Yang, Xi and Xie, Quantao},
    booktitle = {AAAI Conference on Artificial Intelligence},
    volume    = {40},
    number    = {14},
    pages     = {11748--11756},
    year      = {2026},
}

@inproceedings{Tarvainen2017,
  title     = {{Mean Teachers Are Better Role Models: Weight-averaged Consistency Targets Improve Semi-supervised Deep Learning Results}},
  author    = {Tarvainen, Antti and Valpola, Harri},
  booktitle = {Advances in Neural Information Processing Systems},
  volume    = {30},
  year      = {2017}
}

@inproceedings{TentFT,
  title     = {{Tent: Fully Test-Time Adaptation by Entropy Minimization}},
  author    = {Dequan Wang and Evan Shelhamer and Shaoteng Liu and others},
  booktitle = {International Conference on Learning Representations},
  year      = {2021},
}

@inproceedings{TPMOT,
    author    = {Wassim El Ahmar and Angel Sappa and Riad Hammoud},
    title     = {{Thermal Pedestrian Multiple Object Tracking Challenge (TP-MOT)}},
    booktitle = {IEEE/CVF Conference on Computer Vision and Pattern Recognition (CVPR) Workshops},
    year      = {2025},
}

@misc{TrafficCAM,
    author    = {Deng, Zhongying and Cheng, Yanqi and Liu, Lihao and others},
    title     = {{TrafficCAM: A Versatile Dataset for Traffic Flow Segmentation}},
    publisher = {arXiv},
    year      = {2022},
}

@inproceedings{Tran2026,
    author    = {Tran, Duong Nguyen-Ngoc and Pham, Long Hoang and Ho, Quoc Pham-Nam and others},
    title     = {{A Hybrid Data-Centric Framework for Thermal Multiple-Object Tracking with Complex Motion Patterns}},
    booktitle = {IEEE/CVF Conference on Computer Vision and Pattern Recognition (CVPR) Workshops},
    year      = {2026},
    pages     = {7200--7209},
}

@inproceedings{TSBOW, 
    title     = {{TSBOW: Traffic Surveillance Benchmark for Occluded Vehicles Under Various Weather Conditions}}, 
    author    = {Huynh, Ngoc Doan-Minh and Tran, Duong Nguyen-Ngoc and Pham, Long Hoang and others}, 
    booktitle = {AAAI Conference on Artificial Intelligence}, 
    volume    = {40}, 
    number    = {7}, 
    year      = {2026}, 
    pages     = {5239--5247},
}

@inproceedings{YOLO,
    author    = {Joseph Redmon and others},
    title     = {{You Only Look Once: Unified, Real-Time Object Detection}},
    booktitle = {IEEE/CVF Conference on Computer Vision and Pattern Recognition (CVPR)}, 
    year      = {2016},
    pages     = {779--788},
}

@inproceedings{YOLO9000,
    author    = {Joseph Redmon and others},
    title     = {{YOLO9000: Better, Faster, Stronger}},
    booktitle = {IEEE/CVF Conference on Computer Vision and Pattern Recognition (CVPR)}, 
    year      = {2017},
    pages     = {6517--6525},
}

@software{YOLOv5,
    author    = {Glenn Jocher},
    title     = {{YOLOv5 by Ultralytics}},
    year      = {2020},
    version   = {7.0},
    license   = {AGPL-3.0},
}

@inproceedings{YOLOv7,
    author    = {Chien-Yao Wang and others},
    title     = {{YOLOv7: Trainable Bag-of-Freebies Sets New State-of-the-Art for Real-Time Object Detectors}},
    booktitle = {IEEE/CVF Conference on Computer Vision and Pattern Recognition (CVPR)}, 
    year      = {2023},
    pages     = {7464--7475},
}

@misc{YOLO26,
    title         = {{Ultralytics YOLO26: Unified Real-Time End-to-End Vision Models}},
    author        = {Ultralytics},
    year          = {2026},
    eprint        = {2606.03748},
    archivePrefix = {arXiv},
}

@inproceedings{YOLO-World,
    title     = {{YOLO-World: Real-Time Open-Vocabulary Object Detection}},
    author    = {Cheng, Tianheng and Song, Lin and Ge, Yixiao and others},
    booktitle = {IEEE/CVF Conference on Computer Vision and Pattern Recognition (CVPR)},
    year      = {2024},
}

@article{VisDrone,
    title     = {{Detection and Tracking Meet Drones Challenge}},
    author    = {Zhu, Pengfei and Wen, Longyin and Du, Dawei and others},
    journal   = {IEEE Transactions on Pattern Analysis and Machine Intelligence},
    volume    = {44},
    number    = {11},
    pages     = {7380--7399},
    year      = {2021},
}

@inproceedings{VisDrone-MOT2021,
    author    = {Chen, Guanlin and Wang, Wenguan and He, Zhijian and others},
    title     = {{VisDrone-MOT2021: The Vision Meets Drone Multiple Object Tracking Challenge Results}},
    booktitle = {IEEE/CVF International Conference on Computer Vision (ICCV) Workshops},
    year      = {2021},
    pages     = {2839-2846},
}

@article{Zhou2026,
  title     = {{Vision Technologies with Applications in Traffic Surveillance Systems: A Holistic Survey}},
  author    = {Zhou, Wei and Yang, Li and Zhao, Lei and others},
  journal   = {ACM Computing Surveys},
  volume    = {58},
  number    = {2},
  pages     = {1--39},
  year      = {2026},
}
\end{document}